\RequirePackage{amsmath}
\documentclass[runningheads]{llncs}
\usepackage[T1]{fontenc}
\usepackage{graphicx}
\usepackage{amsfonts}
\usepackage{amsmath}
\usepackage{tikz}
\usepackage{pgfplots}
\pgfplotsset{compat=newest,compat/show suggested version=false}
\usepackage{cite}
\usepgfplotslibrary{colorbrewer}
\pgfplotscreateplotcyclelist{foo}{
    {index of colormap=0 of Dark2,mark=*},
    {index of colormap=2 of Dark2,mark=*},
    {index of colormap=3 of Dark2,mark=*},
    {index of colormap=0 of Dark2,mark=triangle*},
    {index of colormap=2 of Dark2,mark=triangle*},
    {index of colormap=3 of Dark2,mark=triangle*},
}
\pgfplotsset{
        cycle list/Dark2,
        cycle multiindex* list={
            foo\nextlist
        },
    }
\usepackage{adjustbox}
\usepackage{subcaption}
\usepackage{multirow}
\usepackage{xcolor}
\definecolor{cvprblue}{rgb}{0.21,0.49,0.74}
\usepackage[pagebackref,breaklinks,colorlinks,citecolor=cvprblue]{hyperref}

\newcommand\nnfootnote[1]{%
  \begin{NoHyper}
  \renewcommand\thefootnote{}\footnote{#1}%
  \addtocounter{footnote}{-1}%
  \end{NoHyper}
}

\begin{document}
\title{
Adapting Large Language Models for Parameter-Efficient Log Anomaly Detection}
\titlerunning{Adapting LLMs for Parameter-Efficient LAD}

%
%
\authorrunning{Y. Lim et al.}
%
\newcommand*{\affaddr}[1]{#1} 
\newcommand*{\affmark}[1][*]{\textsuperscript{#1}}

\author{%
Ying Fu Lim\affmark[*], Jiawen Zhu\affmark[*], Guansong Pang\affmark[\dag]\\
\affaddr{Singapore Management University}
}
%
\maketitle              
%

\begin{abstract}
Log Anomaly Detection (LAD) seeks to identify atypical patterns in log data that are crucial to assessing the security and condition of systems. Although Large Language Models (LLMs) 
 have shown tremendous success in various fields, the use of LLMs in enabling the detection of log anomalies is largely unexplored. This work aims to fill this gap. Due to the prohibitive costs involved in fully fine-tuning LLMs, 
we explore the use of parameter-efficient fine-tuning techniques (PEFTs) 
for adapting LLMs to LAD.
To have an in-depth exploration of the potential of LLM-driven LAD, we present a comprehensive investigation of leveraging two of the most popular PEFTs -- Low-Rank Adaptation (LoRA) and Representation Fine-tuning (ReFT) --
to tap into three prominent LLMs of varying size, including RoBERTa, GPT-2, and Llama-3, for parameter-efficient LAD. Comprehensive experiments on four public log datasets are performed to reveal important insights into effective LLM-driven LAD in several key perspectives, including the efficacy of these PEFT-based LLM-driven LAD methods, their stability, sample efficiency, robustness w.r.t. unstable logs, and cross-dataset generalization. Code is available at \renewcommand\UrlFont{\color{blue}}\url{https://github.com/mala-lab/LogADReft}.

\end{abstract}

\section{Introduction}
Log Anomaly Detection (LAD) plays a vital role in identifying unusual patterns within log data, which are critical to diagnosing system issues and ensuring reliable operations. 
The massive volume and velocity of log data generated by large-scale distributed systems provide valuable insights into real-time operations. However, detecting anomalies based on subtle differences between normal and abnormal log sequences is highly challenging and often exceeds the capacity of human operators to process effectively. To address this, traditional machine learning and deep learning methods have been developed to automatically detect anomalies, thereby reducing the reliance on manual analysis and alleviating the workload of human operators~\cite{du2017deeplog,meng2019loganomaly,zhang2019robust,guo2021logbert}.

\nnfootnote{* Equal contribution}
\nnfootnote{\dag\ Corresponding author: G. Pang (\tt\small gspang@smu.edu.sg)}

Large Language Models (LLMs) have recently emerged as powerful tools for Log Anomaly Detection (LAD) due to their ability to effectively process and comprehend textual data. However, the use of LLMs in enabling the detection of log anomalies is largely unexplored. Recent studies such as \textit{Qi et al.} \cite{qi2023loggptexploringchatgptlogbased} demonstrate the potential of LLMs like ChatGPT in identifying anomalies within log data. However, their performance relies heavily on carefully handcrafted prompts. Besides, fully fine-tuning LLMs for LAD is a resource-intensive process, requiring significant computational power and memory due to their extensive parameter space, limiting their practical adoption in real-world applications.

Existing Parameter-Efficient Fine-Tuning techniques (PEFTs) have emerged as a viable solution to adapt LLMs to various downstream tasks.
Techniques such as Low-Rank Adaptation (LoRA)~\cite{hu2021loralowrankadaptationlarge} and Representation Fine-Tuning~(ReFT) \cite{wu2024reftrepresentationfinetuninglanguage} have demonstrated remarkable efficiency in various resource-constrained environments. LoRA reduces training costs by introducing low-rank matrices to approximate weight updates, whereas ReFT modifies hidden representations with task-specific interventions while preserving the original model weights. Thus, leveraging PEFTs  serves as an important direction for LLM-driven LAD.

In this work we aim to conduct an in-depth exploration of the potential of LLM-driven LAD. To this end, we perform a comprehensive comparison and analysis of the aforementioned two popular PEFT approaches
in unleashing the power of three widely-used LLMs of varying size -- RoBERTa, GPT-2, and Llama-3 -- for LAD. Our goal is not only to achieve efficient and effective LAD through PEFT-based LLM adaptation approaches, but also to evaluate their performance in various other important perspectives, such as model stability, sample efficiency, robustness, and cross-dataset generalization.
In doing so, we reveal several significant insights into LLM-driven LAD. One notable insight is that ReFT-based fine-tuning consistently outperforms LoRA-based approaches in terms of stability, sample efficiency, robustness, and cross-dataset generalization. These results highlight ReFT's potential as a promising direction for LAD applications.

Accordingly, our contributions can be summarized as follows:

\begin{itemize}
\item We perform a pioneering comprehensive evaluation of leveraging different PEFTs to adapt LLMs for LAD.

\item We conduct extensive experiments on two popular PEFTs, LoRA and ReFT, based on three prominent masked/autoregressive LLMs of varying size -- RoBERTa, GPT-2, and Llama-3 -- to achieve LLM-driven, parameter-efficient LAD on four publicly available LAD benchmarks.

\item Our empirical results reveal important insights into five crucial aspects of parameter-efficient LAD, including the efficacy of LLM-based LAD methods, their stability, sample efficiency, robustness to unstable logs, and cross-dataset generalization.
\end{itemize}

\vspace{-0.3cm}
\section{Related Work}
\subsection{Log Anomaly Detection (LAD)}
Numerous methods have been proposed to analyze log data and detect anomalous logs using deep learning. DeepLog \cite{du2017deeplog} is the first LAD method using deep learning, which uses an unsupervised learning-based Long Short-Term Memory (LSTM) to model normal log event sequences and identified anomalies based on deviations. Loganomaly \cite{meng2019loganomaly} extends it by an Attention-based Bi-LSTM
model for unsupervised learning of both frequency and
sequence patterns of log templates. LogRobust \cite{zhang2019robust} and \textit{Lu et al.} \cite{lu2018detecting} perform supervised binary classification on time series of log messages using an Attention-based BiLSTM model and CNN, respectively. Recently, Transformer models have become increasingly popular in LAD due to their better ability to capture long-range dependencies, resulting in stronger transformer-based LAD methods such as LogBERT \cite{guo2021logbert}, FastLogAD \cite{xie2024fastlogendtoendmethodefficiently} and LogELECTRA \cite{yamanaka2024logelectraselfsupervisedanomalydetection}. However, these methods often require extensive training from scratch, which can be difficult to train, and they have poor flexibility in adapting to new log datasets due to a fixed vocabulary space.

\subsection{Tuning Large Language Models(LLMs) for LAD}
Pre-trained LLMs offer significant potential for LAD due to their ability to effectively process and comprehend discrete sequence data. However, fully fine-tuning these large models is resource-intensive. Parameter-Efficient Fine-Tuning (PEFT) methods address this issue by training only a very small fraction of the model's parameters for specific tasks, but their application to LAD remains largely unexplored. Adapter-based approaches like Logformer \cite{guo2024logformerpretraintuningpipeline} introduces a pre-training and adapter-based tuning pipeline to enable the adaptation of transformer-based models to new log datasets. LogBP-LORA \cite{he2023parameter} instead leverages LoRA to turn BERT for LAD. Unlike LogBP-LORA that is focused on small BERT with LoRA, our study offers a systematic evaluation of LoRA and ReFT techniques across multiple large masked and autoregressive LLMs, delivering a comprehensive examination of the capabilities of LLMs in empowering LAD. 
\vspace{-0.3cm}
\section{Methodology} 
\subsection{Problem Statement}
Let $s = \{l_1,l_2,...,l_d\}$ represent a sequence of $d$ log events parsed from raw log messages and grouped together. Let $\mathcal{D}_{train} = \{S_{train}, Y_{train}\}$ denote the training dataset consisting of both normal and anomalous log sequences, where $S_{train} = \{s_i\}_{i=1}^{N}$ represent training log sequences, and $Y_{train} = \{y_i\}_{i=1}^{N} \subset \{0,1\}$ denotes the corresponding ground truth labels, with $y_i=1$ if $s_i$ is abnormal and $y_i=0$ otherwise. Our goal is to train a model using a supervised approach with training data $\mathcal{D}_{train}$. During the inference stage, the trained model is employed to distinguish between normal and anomalous log sequences in the test dataset $D_{\mathrm{test}} = \{S_{test}, Y_{test}\}$.

\vspace{-0.3cm}
\subsection{Parameter-Efficient Log Anomaly Detection}
\subsubsection{Overall Framework.} 
\begin{figure}[t]
    \centering
    \includegraphics[width=0.9\linewidth]{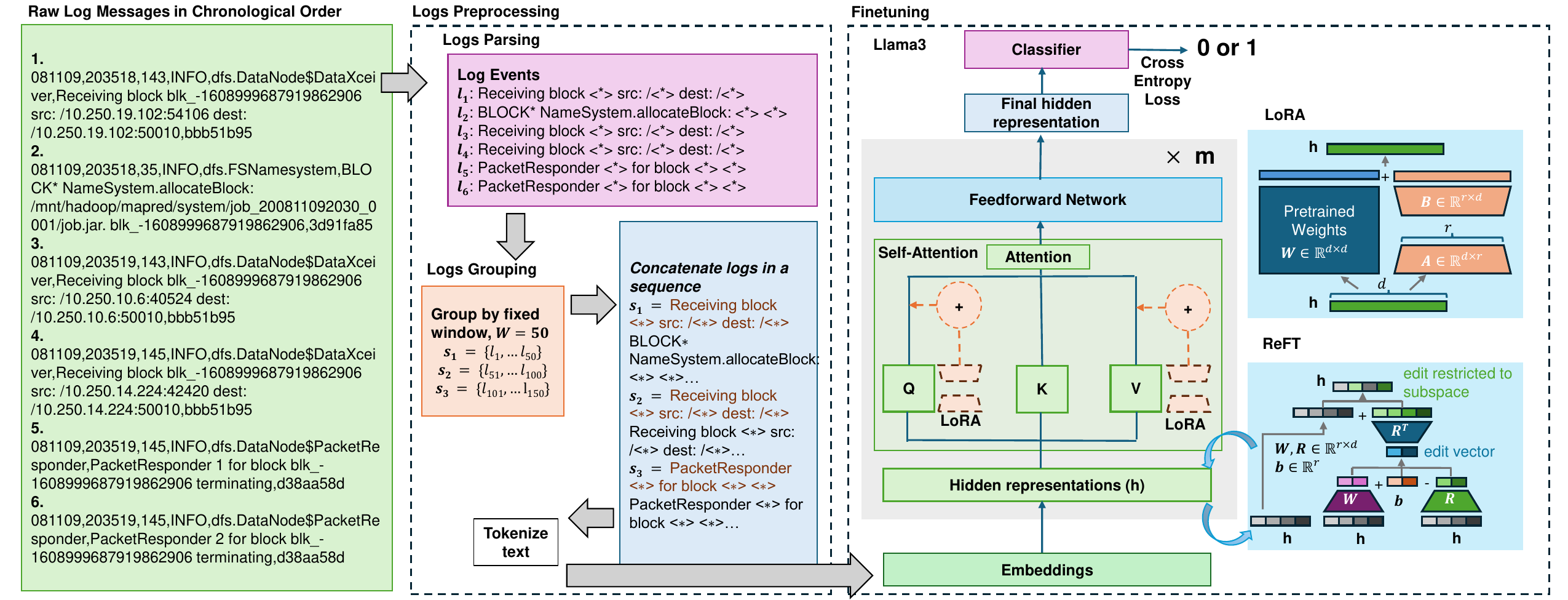}
    \caption{Pipeline of the PEFT-based LLM-driven LAD approaches.}
    \label{pipeline}
\end{figure}

Figure~\ref{pipeline} provides an overview of our approach's framework. In the pre-processing stage, the raw log messages in chronological order are parsed and grouped into log sequences using either a session ID or a fixed sliding window of size 50. These log events within each sequence are concatenated into a single text body, with each event separated by a space. Once grouped, the log sequences are tokenized into token embeddings, which serve as input for the models. 
Our approach utilizes masked or autoregressive LLMs based on the transformer architecture~\cite{vaswani2017attention}. Given an input sequence of tokens $\mathbf{x} = (x_1, ..., x_n)$, the tokens are embedded into an initial set of hidden representations $\mathbf{h}^{(0)} = (\mathbf{h}_1^{(0)}, ..., \mathbf{h}_n^{(0)})$. At the $j^{th}$ layer of the transformer, the hidden representations $\mathbf{h}^{(j)}$ are computed as a function of the preceding layer’s hidden states $\mathbf{h}^{(j-1)}$, with $\mathbf{h} \in \mathbb{R}^{d}$. After passing through $m$ layers, the final hidden representations $\mathbf{h}^{(m)}$ are fed into an anomaly classifier to differentiate between normal and abnormal sequences. For fine-tuning, we employ the LoRA-finetuning method within the self-attention layers of the transformer, while ReFT-finetuning is applied directly to the hidden representations, enhancing the model's ability to capture intricate patterns within the log data.

\vspace{-0.5cm}
\subsubsection{LoRA-finetuned LLMs.}
As illustrated in Figure~\ref{pipeline}, the LoRA method is applied to the query and value matrices within the self-attention layers of the transformer architecture in our experiments. In a linear sub-module of the pre-trained network, with weight parameters $\mathbf{W} \in \mathbb{R}^{d_1 \times k}$ and bias parameters $\mathbf{b} \in \mathbb{R}^{d_2}$, the original transformation is defined as:
\begin{equation}
\mathbf{h}^{(j)}=\mathbf{W}\mathbf{h}^{(j-1)} + \mathbf{b}.
\end{equation}

To incorporate LoRA, this transformation is augmented with an adapter module, which introduces additional parameters $\mathbf{B} \in \mathbb{R}^{d \times r}$, $\mathbf{A} \in \mathbb{R}^{r \times k}$, and a scaling factor $\gamma_{\gamma} \in \mathbb{R}^+$. The LoRA-augmented transformation is defined as:

\begin{equation}
    \mathbf{h^{(j)}_{LoRA}} = (\mathbf{W}+\gamma_{\gamma}\mathbf{BA})\mathbf{h}^{(j-1)} + \mathbf{b}.
\end{equation}

After fine-tuning, the modified weight matrix $(\mathbf{W} + \gamma_{\gamma}\mathbf{BA})$ is stored and used in place of $\mathbf{W}$. The adapter term $\gamma_{\gamma}\mathbf{BA}$ is constrained to have a rank of at most $r$, where $r$ is typically much smaller than both $d_1$ and $k$ (i.e., $r \ll \min(d_1, k)$). This constraint ensures the parameter efficiency of the LoRA module. The scaling factor $\gamma_{\gamma}$ is designed to adjust for the rank effects on the matrix product $\mathbf{BA}$. In rank-stabilized LoRA, $\gamma_{\gamma}$ is defined as:

\begin{equation}
\gamma_{\gamma} = \frac{\alpha}{\sqrt\gamma},
\end{equation}
where $\alpha$ is a hyperparameter that allows fine-tuning of the adjustment to ensure consistent scaling across different rank settings.

\vspace{-0.5cm}
\subsubsection{ReFT-finetuned LLMs.}
In contrast to LoRA, which updates the weights of an additional adapter, ReFT applies interventions directly to the hidden representations before they are passed to the self-attention layer, as shown in Figure~\ref{pipeline}. This intervention is designed to modify the hidden representations in a task-specific manner without altering the pre-trained model’s parameters. The process of ReFT can be formally defined as:

\begin{equation}
    \mathbf{h^{(j)}_{ReFT}=h^{(j-1)}+R^\top(Wh^{(j-1)}+b-Rh^{(j-1)})},
\end{equation}
where $\mathbf{R}$, $\mathbf{W}$, and $\mathbf{b}$ are the learnable parameters involved in the intervention. The term $\mathbf{R} \in \mathbb{R}^{r \times d} $ defines a low-rank projection matrix, while $\mathbf{W}$ and $\mathbf{b}$ represent a linear projection weights and bias, respectively. The matrix $\mathbf{R} $ has $r$ orthonormal rows, and the rank $r$ is typically smaller than the dimensionality $d$ of the hidden state ($r \leq d$).

\vspace{-0.5cm}
\subsubsection{Training and Inference in PEFT-based LAD.}
After passing through $m$ layers of fine-tuning, the model extracts a specific token representation from the final layer, denoted as $\mathbf{h}^{(m)}$. For masked LLMs (e.g., RoBERTa), the representation of the special token $[CLS]$ is used. In contrast, for autoregressive LLMs (e.g., GPT-2), the representation of the final token in the sequence is utilized. This token embedding, represented as $\mathbf{h}_t^{(m)}$ (where $t=1$ for masked LLMs and $t=n$ for autoregressive LLMs), is subsequently passed to a classifier $\Psi(\cdot)$, parameterized by $\theta$, which outputs a binary prediction $\Psi(\mathbf{h}_t^{(m)}) \in \{0, 1\}$ to indicate whether the log sequence is normal or anomalous, with larger score indicates higher probability of $x$ to be abnormal.

The training objective is to optimize the model for accurate classification by minimizing loss for the target class $y$ given the input $x$. This objective is formalized as:

\begin{equation}
\mathcal{L} = \sum_{i=1}^{N}\mathcal{L}_{ce}\left(\Psi(\mathbf{h}_t^{(m)}), y\right),
\label{learning}
\end{equation}
where $\mathcal{L}_{ce}$ is the cross-entropy loss, and $N$ is the total number of training data.

During inference, a test image $x^\prime$ is processed through the trained PEFT-based LAD model to obtain the token representation from the final layer, denoted as $\mathbf{h^\prime}^{(m)}$. The final anomaly score is then computed using $\Psi(\mathbf{h^\prime}^{(m)})$.

\section{Experiments}
\noindent\textbf{Datasets.} 
To evaluate the efficiency of LLM-based parameter-efficient techniques in Log Anomaly Detection, we conducted a comprehensive set of experiments on four publicly available real-world log datasets, including HDFS~\cite{10.1145/1629575.1629587}, BGL~\cite{Oliner2007WhatSS}, Spirit~\cite{Oliner2007WhatSS}, and Thunderbird~\cite{Oliner2007WhatSS}. 
Following previous work~\cite{le2022log}, we leverage the front 80\% (according to the timestamps of logs) as the training data, and the rest 20\% as the testing data.

\noindent\textbf{Competing Methods and Evaluation Metrics.}
We compare our parameter-efficient methods against several state-of-the-art approaches, including three unsupervised methods (LogBERT~\cite{guo2021logbert}, DeepLog~\cite{du2017deeplog}, LogAnomaly~\cite{meng2019loganomaly}) and two supervised methods(CNN~\cite{lu2018detecting}, LogRobust~\cite{zhang2019robust}). 
For evaluation metrics, we adopt the widely used precision, recall, and F1-score to assess the performance of all methods.

\begin{table}[t]
    \caption{Results of log anomaly detection performance of the various models on various datasets}
    \centering
    \begin{subtable}{0.4\textwidth}
    \centering
    \caption{BGL \label{BGL results}}
    \adjustbox{max width=\textwidth}{
    \begin{tabular}{|l|c|c|c|}
        \hline
        \textbf{Methods} & \textbf{F1} & \textbf{Precision} & \textbf{Recall} \\ \hline
        \textbf{RoBERTa-ReFT} & 0.9558 & 0.9693 & 0.9433 \\ 
        \textbf{GPT2-ReFT} & 0.8919 & 0.8541 & 0.9417 \\ 
        \textbf{Llama3-ReFT} & \textbf{0.9690} & \textbf{0.9738} & 0.9643 \\ 
        \textbf{RoBERTa-LoRA} & 0.9299 & 0.9703 & 0.8927 \\ 
        \textbf{GPT2-LoRA} & 0.9136 & 0.9210 & 0.9064 \\ 
        \textbf{Llama3 -LoRA} & 0.9485 & 0.9629 & 0.9345 \\ 
        \textbf{LogRobust} & 0.9685 & 0.9617 & \textbf{0.9755} \\ 
        \textbf{CNN} & 0.7597 & 0.6293 & 0.9582 \\ 
        \textbf{LogBERT} & 0.2496 & 0.1433 & 0.9683 \\ 
        \textbf{DeepLog} & 0.2852 & 0.167 & 0.9748 \\ 
        \textbf{LogAnomaly} & 0.2848 & 0.1668 & 0.9726 \\ 

        \hline
    \end{tabular}
    }
    \end{subtable}
    \hfill
    \begin{subtable}{0.4\textwidth}
    \centering
    \caption{HDFS \label{HDFS results}}
    \adjustbox{max width=\textwidth}{
    \begin{tabular}{|l|c|c|c|}
        \hline
        \textbf{Methods} & \textbf{F1} & \textbf{Precision} & \textbf{Recall} \\ \hline
        \textbf{RoBERTa-ReFT} & 0.9965 & \textbf{0.9972} & 0.9958 \\ 
        \textbf{GPT2-ReFT} & 0.9971 & 0.9964 & 0.9978 \\ 
        \textbf{Llama3-ReFT} & 0.9917 & 0.9956 & 0.9878 \\ 
        \textbf{RoBERTa-LoRA} & 0.9941 & 0.9950 & 0.9932 \\ 
        \textbf{GPT2-LoRA} & 0.9951 & 0.9947 & 0.9956 \\ 
        \textbf{Llama3 -LoRA} & \textbf{0.9972} & 0.9950 & \textbf{0.9994} \\ 
        \textbf{LogRobust} & \textbf{0.9972} & 0.9952 & 0.9991 \\ 
        \textbf{CNN} & 0.9797 & 0.9848 & 0.9745 \\ 
        \textbf{LogBERT} & 0.7736 & 0.9912 & 0.6343 \\ 
        \textbf{DeepLog} & 0.5765 & 0.5047 & 0.6721 \\ 
        \textbf{LogAnomaly} & 0.8156 & 0.8877 & 0.7543 \\ 
  
        \hline
    \end{tabular}
    }
    \end{subtable}
    \vfill
    \begin{subtable}{0.4\textwidth}
    \centering
    \caption{Spirit \label{Spirit results}}
    \adjustbox{max width=\textwidth}{
    \begin{tabular}{|l|c|c|c|}
        \hline
        \textbf{Methods} & \textbf{F1} & \textbf{Precision } & \textbf{Recall} \\ \hline
        \textbf{RoBERTa-ReFT} & 0.7940 & 0.9164 & 0.7272 \\ 
        \textbf{GPT2-ReFT} & 0.8499 & 0.8874 & 0.8181 \\ 
        \textbf{Llama3-ReFT} & 0.9262 & \textbf{0.9642} & 0.8939 \\ 
        \textbf{RoBERTa-LoRA} & 0.7719 & 0.9167 & 0.6667 \\ 
        \textbf{GPT2-LoRA} & 0.5769 & 0.7895 & 0.4545 \\ 
        \textbf{Llama3 -LoRA} & \textbf{0.9846} & 1.0000 & \textbf{0.9697} \\ 
        \textbf{LogRobust} & 0.8955 & 0.8824 & 0.9091 \\ 
        \textbf{CNN} & 0.8621 & 1.0000 & 0.7576 \\ 
        \textbf{LogBERT} & 0.0488 & 0.0253 & 0.6667 \\ 
        \textbf{DeepLog} & 0.0537 & 0.0276 & \textbf{0.9697} \\ 
        \textbf{LogAnomaly} & 0.0537 & 0.0276 & \textbf{0.9697} \\ 
        \hline
    \end{tabular}
    }
    \end{subtable}
    \hfill
    \begin{subtable}{0.4\textwidth}
    \centering
    \caption{Thunderbird \label{Thunderbird results}}
    \adjustbox{max width=\textwidth}{
    \begin{tabular}{|l|c|c|c|}
        \hline
        \textbf{Methods} & \textbf{F1} & \textbf{Precision } & \textbf{Recall} \\ \hline
        \textbf{RoBERTa-ReFT} & 0.8138 & 0.9000 & 0.7593 \\ 
        \textbf{GPT2-ReFT} & 0.8604 & 0.8421 & 0.8809 \\ 
        \textbf{Llama3-ReFT} & \textbf{0.9149} & 0.9459 & 0.8817 \\ 
        \textbf{RoBERTa-LoRA} & 0.6838 & 0.8348 & 0.5791 \\ 
        \textbf{GPT2-LoRA} & 0.7482 & 0.7660 & 0.7312 \\ 
        \textbf{Llama3 -LoRA} & 0.9066 & \textbf{0.9929} & 0.8340 \\ 
        \textbf{LogRobust} & 0.8584 & 0.9475 & 0.7846 \\ 
        \textbf{CNN} & 0.7458 & 0.7960 & 0.7016 \\ 
        \textbf{LogBERT} & 0.4000 & 0.4500 & 0.3600 \\ 
        \textbf{DeepLog} & 0.8561 & 0.8121 & \textbf{0.9051} \\ 
        \textbf{LogAnomaly} & 0.8561 & 0.8121 & \textbf{0.9051} \\ 
        \hline
    \end{tabular}
    }
    \end{subtable}
\label{main_results}
\vspace{-0.5cm}
\end{table}

\noindent\textbf{Implementation Details.} \label{param-setting}
To comprehensively analyze the efficiency of parameter-efficient methods, we selected three prominent LLMs as backbones: RoBERTa~\cite{DBLP:journals/corr/abs-1907-11692} with 125M parameters, GPT2~\cite{Radford2019LanguageMA} with 117M parameters and Llama3~\cite{dubey2024llama3herdmodels} with 8B parameters. For LoRA-finetuned LLMs, we implemented LoRA using rank-stabilized LoRA~\cite{kalajdzievski2023rankstabilizationscalingfactor} from the HuggingFace library~\cite{wolf-etal-2020-transformers}. By default, we set the rank to 128 and $\alpha$ to 256 across all models. LoRA was applied across all layers, specifically targeting the query, key, and value weights for GPT2, and query and value weights for RoBERTa and Llama3 within the attention module of the transformer. For ReFT-finetuned LLMs, we used the LoReFT implementation provided by \textit{Wu et al.}~\cite{wu2024reftrepresentationfinetuninglanguage}. The rank is set to 8 and we intervene only at the first position (prefix) for RoBERTa and the last position (suffix) for GPT-2 and Llama3. By default, we use the AdamW optimizer with an initial learning rate of $1e^{-4}$. The batch size is set to 32 for both RoBERTa and GPT-2 and 8 for Llama-3. For the implementation of baseline models, we adhere to the code and settings established by \textit{Le et al.} \cite{le2022log}. 

\vspace{-0.3cm}
\subsection{Efficacy of PEFT-based LAD}
\noindent\textbf{Detection Accuracy.}
We evaluated our parameter-efficient model against state-of-the-art baselines on the BGL, HDFS, Spirit, and Thunderbird datasets, with the results summarized in Table~\ref{main_results}. Compared to unsupervised approaches (LogBERT, DeepLog, and LogAnomaly) and deep network based methods (CNN and LogRobust), LLM-driven fine-tuning models consistently demonstrated superior performance. 
Among the models, PEFT-based Llama-3 showed the best overall performance across all datasets. This is likely due to Llama-3’s larger parameter size and having trained on a larger corpus, enabling it to identify and capture complex patterns more effectively. Furthermore, ReFT-finetuned LLMs outperformed LoRA-finetuned LLMs in 9 out of 12 comparisons, demonstrating the superior performance of ReFT.

\begin{figure*}[t]
    \centering
    \includegraphics[width=0.75\linewidth]{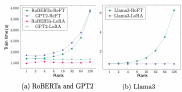}
    \caption{Training time per epoch against rank for various fine-tuned LLMs on BGL dataset.}
    \label{fig:bgl-train-time-plot}
    \vspace{-0.5cm}
\end{figure*}

\vspace{-0.5cm}
\subsubsection{Time Efficiency.}
To compare the model complexity of our parameter-efficient techniques, we measured the training time per epoch for ReFT- and LoRA-finetuned LLMs on the BGL dataset. The results are illustrated in Figure~\ref{fig:bgl-train-time-plot}(a) and Figure~\ref{fig:bgl-train-time-plot}(b), respectively. We observe that the training time for ReFT-finetuned LLMs increases exponentially as the rank and the number of intervenable parameters grow. In contrast, the training time for the LoRA-finetuned LLMs remains steady, regardless of changes in rank or trainable parameters. 

Despite ReFT's higher time consumption, it offers substantial performance improvements over LoRA. To balance training time and performance gains, we set the default rank to 8 for our experiments. At this rank, Llama3-ReFT requires roughly 40\% more training time compared to LoRA (9,943 seconds vs. 7,064 seconds), adding an additional 2.4 hours when fine-tuning for the standard three epochs. This makes ReFT a viable and effective approach in log anomaly detection despite its higher time requirement.

\begin{figure*}[t]
    \centering
    \includegraphics[width=0.75\linewidth]{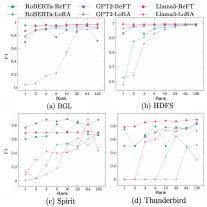}
    \caption{F1 score against rank for various finetuned LLMs on various datasets.}
    \label{rank}
    \vspace{-0.5cm}
\end{figure*}

\vspace{-0.5cm}
\subsubsection{Effectiveness of Using Varying Ranks.}
To evaluate the impact of rank on the efficiency of the fine-tuned methods, we compare the F1 scores of various fine-tuned LLMs across different rank values on four datasets. The results are shown in Figure~\ref{rank}. Following the experimental settings described in Section \ref{param-setting}, we vary the rank as powers of two, from 1 to 128. In general, ReFT-finetuned LLMs consistently outperform LoRA-finetuned LLMs across most rank values and datasets. Notably, ReFT achieves strong performance even at lower ranks, whereas LoRA requires significantly higher ranks to reach comparable results, particularly on more complex datasets with larger sample sizes and diverse patterns, such as Thunderbird.  This can be attributed to ReFT’s parameter efficiency, enabling it to achieve similar or better performance with fewer parameters~\cite{wu2024reftrepresentationfinetuninglanguage}. 

Both ReFT and LoRA generally show improved performance as the rank increases, likely due to the model's enhanced ability to capture more complex patterns in the log data. Notably, larger LLMs tend to require smaller rank values to achieve optimal results in LoRA-based methods. However, performance begins to decline beyond a certain rank, suggesting possible overfitting, where the model starts capturing specific patterns in the training data that do not generalize well to unseen data.

\begin{figure*}[t]
    \centering
    \includegraphics[width=0.75\linewidth]{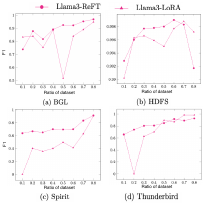}
    \caption{F1 score against ratio of the dataset for Llama3 on various dataset.}
    \label{training_data}
    \vspace{-0.5cm}
\end{figure*}

\vspace{-0.5cm}
\subsubsection{Scalability and Resource Trade-offs.}
Overall, PEFT-based LLMs offer a practical and scalable solution for LAD, striking a balance between model efficiency and detection performance. While ReFT may require longer training times and increased memory usage at higher ranks, it provides significant advantages in anomaly detection accuracy, making it a strong candidate for high-stakes anomaly detection tasks. Meanwhile, LoRA’s stable training time and lower memory footprint make it a more efficient choice for resource-constrained environments. These trade-offs should be carefully considered when selecting a fine-tuning approach for real-world LAD deployments.

\vspace{-0.3cm}
\subsection{Sample Efficiency, Robustness, and Generalization}
\noindent\textbf{Sample Efficiency.}
As the pretrained LLMs were trained on a large corpus of data, the finetuning process generally requires a small amount of data to achieve good performance. To explore this, we investigate how the performance of LoRA- and ReFT-finetuned Llama3 change when we adjust the amount of train data available. Specifically, we use the same settings described in Section \ref{param-setting} implementation details with a fixed epoch of 3. The test dataset is fixed at the last 20\% of the data, while the percentage of the training set of the dataset ranges from 10\% to 80\%. The datasets are arranged in chronological order. Figure~\ref{training_data} presents the F1 scores for various fine-tuning approaches on the BGL, HDFS, Spirit, and Thunderbird datasets.

Across all datasets, Llama3-ReFT generally outperforms Llama3-LoRA and requires less training data to achieve strong performance. This can be attributed to ReFT’s greater parameter efficiency compared to LoRA. As expected, performance improves for both approaches as the training data ratio increases, due to the model's access to more data for learning.

\begin{figure}[h]
    \centering
    \begin{minipage}{0.4\textwidth}
    \centering
    \includegraphics[width=\linewidth]{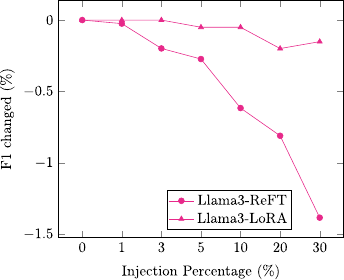}
    \caption{F1 change (\%) against injection percentage (\%) for Llama3 on HDFS dataset}
    \label{fig:hdfs-unstable-plot}
    \end{minipage}
    \hfill
    \begin{minipage}{0.4\textwidth}
        \centering
        \captionof{table}{F1 score of Llama3 finetuned on Spirit dataset across various dataset.}
        \adjustbox{max width=\textwidth}{
        \begin{tabular}{|l|c|c|}
        \hline
        \multirow{2}{*}{\textbf{Test Dataset}} & \multicolumn{2}{c|}{\textbf{F1-score}} \\ 
        \cline{2-3} & \textbf{Llama3-ReFT} & \textbf{Llama3-LoRA} \\ \hline
        \textbf{BGL} & 0.4823 & 0.0072 \\ 
        \textbf{HDFS} & 0.4925 & - \\ 
        \textbf{Thunderbird} & 0.5864 & 0.0568 \\ 
        \textbf{Spirit} & 0.9192 & 0.8849 \\ 
        \hline
        \end{tabular}
        }
        \label{zero_shot}
    \end{minipage}
    \vspace{-0.5cm}
\end{figure}

\vspace{-0.5cm}
\subsubsection{Robustness to Unstable Logs.}
Logs are generated by a variety of applications, and their vocabulary can evolve over time due to feature changes, leading to instability in log data. To assess the impact of such unstable logs on the performance of LoRA- and ReFT-finetuned LLMs, we conduct experiments where the vocabulary of log events in the test dataset is altered. Particularly, we replace the top 10 action words in the log events of the test dataset with the top 1-3 synonyms from WordNet~\cite{miller-1994-wordnet}, to simulate the unstable logs as vocabulary changes. The test dataset consists of 50,000 normal logs and 1,000 anomalous logs,
and we inject unstable logs into the test dataset at rates of 1\%, 2\%, 3\%, 5\%, 10\%, 20\%, and 30\%. All other settings are consistent with those described in Section \ref{param-setting} implementation details, except that the training epoch is set to 1.

As shown in Figure \ref{fig:hdfs-unstable-plot}, as the injection percentage of unstable logs increases, the performance of Llama3-ReFT slightly decreases, while the F1 score of Llama3-LoRA remains relatively stable. This indicates that the ReFT approach is more sensitive to unstable logs compared to the LoRA approach. However, the performance drop for Llama3-ReFT is minimal -- only around 1.4\% in F1 score at a 30\% unstable log injection. This suggests that the impact of unstable logs on Llama3-ReFT’s performance is limited. Overall, both fine-tuning approaches show a reasonable degree of robustness when the vocabulary of the logs changes.

\vspace{-0.5cm}
\subsubsection{Cross-dataset Generalization.}
To investigate the generalization capability of finetuning methods in log anomaly detection, Llama-3 was finetuned using the Spirit training dataset, and the resulting finetuned model was subsequently applied to predict anomalies in the BGL, HDFS, and Thunderbird test datasets. As shown in Table~\ref{zero_shot}, the ReFT-based Llama-3 model significantly outperforms the LoRA-based model, achieving an F1 score of approximately 0.5 across the test datasets. In contrast, the F1 score of the LoRA-based model remains close to 0 across these datasets (the result of Llama3-LoRA on HDFS is empty due to it only output 1 for all test samples). This is likely due to the overfitting of LoRA finetuned Llama3 to the train dataset so that unable to perform well with test datasets. Although ReFT has better performance than LoRA, the F1 score of 0.5 means that it does not effectively identify anomalous logs for the dataset it has not seen. These findings confirm that ReFT typically outperforms LoRA while being more parameter efficient in the field of LAD, in agreement with prior research.

\vspace{-0.3cm}
\section{Conclusions}
In this paper, we provide a thorough exploration of adapting large language models (LLMs) for log anomaly detection (LAD) through parameter-efficient fine-tuning techniques (PEFTs). By leveraging LoRA and ReFT variants within three well-established LLMs—RoBERTa, GPT-2, and Llama-3—we demonstrate the potential of LLM-driven, parameter-efficient LAD. Our extensive experiments on four public log datasets not only validate the effectiveness of these methods but also offer valuable insights into five critical aspects of parameter-efficient LAD, including the effectiveness of LLM-driven methods, their stability, sample efficiency, robustness to unstable logs, and cross-dataset generalization.

\vspace{-0.3cm}
\section*{Acknowledgments}
This work is supported by the Ministry of Education, Singapore under its Tier-1 Academic Research Fund (24-SIS-SMU-008) and the Lee Kong Chian Fellowship.

\bibliography{bibliography}
\bibliographystyle{splncs04}

\end{document}